# Corpus of Chinese Dynastic Histories: Gender Analysis over Two Millennia

## Sergey Zinin, Yang Xu

University of Massachusetts Amherst, University of Toronto
Warring States Workshop, Amherst, Massachusetts, USA, Department of Computer Science, Toronto, Canada
szinin@research.umass.edu, yangxu@cs.toronto.edu

## Abstract

Chinese dynastic histories form a large continuous linguistic space of approximately 2000 years, from the 3[rd] century BCE to the 18[th] century CE. The histories are documented in Classical (Literary) Chinese in a corpus of over 20 million characters, suitable for the computational analysis of historical lexicon and semantic change. However, there is no freely available open-source corpus of these histories, making Classical Chinese low-resource. This project introduces a new open-source corpus of twenty-four dynastic histories covered by Creative Commons license. An original list of Classical Chinese gender-specific terms was developed as a case study for analyzing the historical linguistic use of male and female terms. The study demonstrates considerable stability in the usage of these terms, with dominance of male terms. Exploration of word meanings uses keyword analysis of focus corpora created for gender-specific terms. This method yields meaningful semantic representations that can be used for future studies of diachronic semantics.

## 1. Introduction

The renowned sinologist Homer L. Dubs noted, "the world's greatest repository of historical information is to be found in the twenty-five officially approved Chinese standard histories" (Dubs, 1946). This happened due to centrality of historiography toward the traditional Chinese culture. One result of this fact is the availability of a large resource for corpus linguistic studies of Classical Chinese. Unlike many of the early Chinese literature, the Twenty-Four Dynastic Histories [1] have solid textological provenance; they were written under rigorous stylistic requirements in consistent Classical Chinese over a period of more than 2000 years (from the 3[rd] century BCE to the 18[th] century CE), making them one of the longest diachronic linguistic repositories.

While some parts of these histories (e.g., chronicles *benji*) are considered very formal, the main body of histories consists of "biographies" (*liezhuan*). These biographical narratives present the life of Chinese society in different periods. Although philosophical treatises of the Warring States and Early Imperial periods have invited research among computational linguists, few can compare with the dynastic histories as a linguistic resource. The size of this resource [2] and its diachronic scope make it ideal for corpus linguistic studies of Classical Chinese.

The significance of the Twenty-Four Histories as the major resource for computational linguistics has always been understood in the Chinese academy. The earliest and one of the most authoritative online corpora of Classical Chinese, the Academia Sinica's *Scripta Sinica* (see Scripta Sinica), was built on "Full Text Chinese documents" database project, which itself started as digitization of the dynastic histories [3]. However, despite having been digitized and placed online, this resource is not available to the community at large. As Li et al. (2012) wrote, "in recent years, the Academia Sinica has constructed a corpus of Pre-Qin Chinese containing 20 classical literatures. However, this important resource only supplies online queries, and has not been used to get a statistical overview of the Pre-Qin vocabulary by the developers." This is still true. Most researchers in the Classical Chinese field do not provide access to the corpora they worked on, and access to academic and commercial resources is restricted or prohibited [4]. Only in 2014, Song and Xia (2014) presented free open-source corpus of the Huainanzi—probably, the first such corpus available.

There has been a recent growth in open linguistic diachronic resources, e.g., the HistWords project by W.L. Hamilton, J. Leskovec and D. Jurafsky (see HistWords). These authors note, that, in their existing resource, "Chinese lacks sufficient historical data for this task, as only years 1950-1999 are usable" (Hamilton et al., 2016). This project aims to improve the availability of open-source corpora of Classical Chinese by offering free corpora that can be downloaded and used.

One source of digital Classical Chinese data is the Chinese Wikisource (see Wikisource). The Wikisource

---

[1] In the current project, the term Twenty-Four Histories is used, as it is more traditional. If the history of Qing dynasty, the Qingshi, added, we could talk about the Twenty-Five Histories, as in the quotation from Dubs.

[2] There are various estimates, depending on the source: over 20 million tokens, by Dubs' estimate (Dubs, 1946), or rather 39 million tokens (Lee and Chen, 1997), or 31 million tokens (Huang and Wu). This project presents a corpus with over 23 million characters (see Section 2).

[3] "The original project began under the name "Computerization of Historical Documents" in 1984 when researchers from both the Institute of History and Philology and the Computing Center at Academia Sinica worked together to key in the "Monographs on Economy" from the dynastic histories. In 1986, the project was expanded to include the entire twenty-five dynastic histories. In June 1990, the computerization of the full text of the dynastic histories was completed with the exception of the charts." This part of the project took six years and cost more than NT$40,000,000 (approximately US$1,400,000). The "database of the twenty-five dynastic histories" is the first and largest segment of the Full Text Project" (Lee and Chen, 1997).

[4] For example, Li's own project group that worked on Ancient Chinese Corpus (ACC) V1.0 since 2009 at the Nanjing Normal University, as far as these authors found, only released one classic (the Zuozhuan) on the commercial resource of LDC in 2017.

contains the full text of the Twenty-Four Histories under the Creative Commons license, i.e., they could be freely used, re-distributed, and modified. However, there are two limitations: the philological provenance and the current format. It should be noted that, unlike many other online resources of Classical Chinese, the Wikisource is openly editable. The authors presume that the philological quality of Wikisource sources is sufficient for corpus analysis while challenging for exact textological studies. The other drawback is its online format. Copying the textual data, cleaning them up, and reformatting take considerable time. This project introduces the Corpus of Chinese Dynastic Histories (CCDH, 2019), with a total of over 23 million characters[5].

The authors hope that the CCDH corpus will alleviate the lack of diachronic resources of Classical Chinese. Diachronic, or historical, linguistics has been developing in the Western humanities since the 18th century because of growing availability of large diachronic digital corpora. In particular, new computational linguistic methods have been developed to analyze semantic change (see Tang, 2018, for an up-to-date review).

The dynastic histories contain a vast amount of information on the traditional Chinese society and Classical Chinese language, especially in *liezhuan* (biographies) sections, which makes the bulk of the corpus. This project focuses on the subject of gender analysis of the dynastic histories. In modern linguistics, Chinese language is generally considered "genderless," i.e., it lacks not only grammatical but also natural gender category (Stahlberg et al., 2007). Also, modern gender analysis such as "gender classification" and "gender identification" are not applicable to the dynastic histories because they are all known to be written by men and from a masculine discourse position.

Farris (1988) is a pioneer in gender analysis of Modern Chinese, and her approach of studying gender through covert and marked terms is still significant. Working in a pre-digital age, she created the first lists of male and female gendered terms (or gender-specific terms), and this approach forms the basis of this study; gender-specific terms of such categories as sex, kin, and official ranks were identified based on the vocabulary of the histories, and evidences of their semantic contexts were explored.

Aside from Farris' lists (created for Modern Chinese), the list of gender-specific terms created in this study for the dynastic histories is probably the first and largest of its kind. It should be noted that it is not a comprehensive list of all such terms for every historical period. This study has been, from the beginning, a diachronic investigation, and for over 2000 years, many terms, especially the official ranks, which were introduced in the Shiji, were extinguished by the Mingshi time[6]. It was challenging to create an exhaustive list of gender-specific terms, which are present in all dynastic histories, from the Shiji (period of Classical Chinese) to the Mingshi periods (period of early Modern Chinese), and most of them are still present in Modern Chinese. Considering the novelty of this task, the analysis of the usage of these terms has been limited by the analysis of their context within the context windows of sentence and paragraph structures. The idea was to try to establish the existence of special relations of certain context terms with male and female terms.

Therefore, this study did not directly address the issue of semantic change. Rather it offered an initial exploration of the contextual (or semantic) environment of individual terms. A special focus sub-corpus of all sentences and paragraphs, where the term was entered, was created for each term. Previous researchers have implemented this method of term semantic analysis — e.g., Lau and Cook (2012)—for the identification of novel senses of words. However, Lau and Cook used topic modeling with Latent Dirichlet Allocation and Hierarchical Dirichlet Process (LDA/HDP) to extract senses of words. This method did not perform well in the current study of Classical Chinese corpus, where there is no word mark-up. Therefore, the keyword analysis (KA) method also has been implemented, and it has yielded meaningful results.

## 2. Corpus Description

### 2.1. Dynastic Histories in the Corpus

The Chinese language has evolved in roughly three stages: Old Chinese, Medieval Chinese, and Modern Chinese (Norman, 1998). Classical Chinese could be considered as a written form of Old Chinese, which formed in the past three centuries BCE (Dong, 2014). In post-Han period, it could be referred to as Literary Chinese. It is often perceived that Classical or Literary Chinese practically has not changed since then, unlike spoken language. Researchers agree that there should have been some intercommunication between the current spoken language and Classical Chinese, although this area has not yet been considerably explored. The processed text of histories, with dynastic periods and basic statistics, are presented in Table 1, and the Corpus of Chinese Dynastic Histories (CCDH) can be found at https://osf.io/tp729/.

### 2.2. Corpus Creation and Composition

The text of the twenty-four histories was taken from Wikisource and processed to remove all formatting, except divisions by chapters (*juan*), paragraphs, and sentences. Although the text was already present online, converting it into a corpus that could be used for various Classical Chinese research required dedicated work, which makes this corpus a unique contribution. The chapter numbers were entered in the form of *001* (for *juan* 1) on a separate line. The text files, in UTF8 coding, could be found on the project GitHub site; they have names such as "01_shiji_full.txt" (for the Shiji)[7].

---

[5] This confirms the conservative evaluation by Dubs (Dubs, 1946); however, with punctuation marks and white spaces the volume of CCDH is over 27 million tokens, which is close to (Huang and Wu, 2018) evaluation (the latter study also includes the large Qing dynasty history).

[6] See Table 1 for the dynasties' creation times.

[7] The description of file content of the OSF site is contained in the README file on it. Specific file names will be omitted farther on.

These files were further parsed and processed to create index files, where each character is placed on a separate line in a related text file

| Name | Dynasty | Period | Chars | Types |
|------|---------|--------|-------|-------|
| Shiji | To Han | To 95 bce | 577256 | 5045 |
| Hanshu | Han | 206 bce – 24 ce | 773741 | 5906 |
| Houhanshu | Houhan | 25-220 | 690771 | 5553 |
| Sanguozhi | Wei, Wu, Shu | 221-280 | 384155 | 4489 |
| Jinshu | Jin | 265-420 | 1149450 | 5794 |
| Songshu | Liu-Song | 420-479 | 800235 | 5825 |
| Nanqishu | Nan Qi | 479-502 | 296729 | 4859 |
| Liangshu | Liang | 502-557 | 293085 | 4915 |
| Chenshu | Chen | 557-589 | 163125 | 3970 |
| Weishu | Wei | 386-550 | 965445 | 5325 |
| Beiqishu | Bei Qi | 550-577 | 213172 | 3992 |
| Zhoushu | Zhou | 535-581 | 260865 | 4136 |
| Suishu | Sui | 581-618 | 693690 | 5544 |
| Nanshi | Nan chao | 420-589 | 675661 | 5160 |
| Beishi | Bei chao | 386-618 | 1103684 | 5515 |
| Jiutangshu | Tang | 618-906 | 1984156 | 6382 |
| Xintangshu | Tang | 618-907 | 1769453 | 6838 |
| Jiuwudaishi | Wu dai | 907-960 | 605041 | 4661 |
| Xinwudaishi | Wu dai | 907-960 | 290748 | 3922 |
| Songshi | Song | 960-1279 | 3995199 | 11254 |
| Liaoshi | Liao | 907-1125 | 300866 | 3994 |
| Jinshi | Jin | 1115-1234 | 940129 | 5102 |
| Yuanshi | Yuan | 1271-1368 | 1591729 | 5744 |
| Mingshi | Ming | 1368-1644 | 2828640 | 7407 |
| Total | | | 23347014 | 15071 |

Table 1: Text statistics for the twenty-four dynastic histories (length in characters with punctuations removed).

and is provided with history, chapter, paragraph, sentence numbers, and position in the sentence, e.g., a line like:

1,113,10,1,22,主

means that this character is found in the History Number 1 (the Shiji), Chapter 113, 10[th] paragraph in this chapter, first sentence in this paragraph, and 22[nd] position in the sentence. All further experiments were conducted using these index files, not the original text files.

## 3. Gender Terms

Due to the inclusion of the "biographies" section, the dynastic histories in the corpus contain not just purely historical data but also information on many aspects of everyday life in China, including family stories, where data on gender relations could be found. Therefore, it seems natural to do a gender analysis and consider what information about gendered terms could be extracted.

As appeared in the bibliography collected by Marjorie Chan (see Chan), most linguistic gender studies of Chinese could be classified through analysis of the history of pronouns, gender identification, and special women's language. There are limited studies of semantic gender analysis of Classical Chinese. One of the established methods of gender analysis used lists of gender-specific terms (Crawford et al., 2004). As mentioned, Chinese is genderless; therefore, it appears that the most appropriate way to analyze gender in a Classical Chinese corpus is through semantic analysis of gender-specific words, i.e., such words that refer only to either males or females. Such words could be found among family, kinship, and professional terms. Farris (1988) created several similar lists, but her lists are based on Modern Chinese. However, gender terms in Modern Chinese cannot form the immediate basis of investigation into a corpus of Classical Chinese because of language change. Creating such a gender-specific list of terms for Classical Chinese is itself challenging. Even more challenging is that such a list should be applicable to a diachronic corpus of almost 2-millennia scale, as official titles (an equivalent of occupational terms for dynastic histories), male and female, often do not last longer than one or two dynasties[8]. To tackle this issue, the authors have identified (via unigrams, bigrams, and trigrams) potential words that should be on the list and are present in most critical histories in the current corpus and extract their dictionary values, using CC-CEDICT dictionary project[9]. The entries on the gender-specific list include full character form, simplified form, pinyin Romanization, and English translations by CC-CEDICT. See a sample in Table 2. There are 81 male and 31 female terms[10]. The difference in numbers was caused by 1) the nature of historical

---

[8] Hucker's indispensable dictionary of the official titles in imperial China (Hucker, 1985) had been consulted in the process of this work; however, it could not offer a ready-made solution.

[9] CC-CEDICT dictionary, which is used by many Chinese language-related projects, e.g., by UNIHAN database, is also covered by Creative Commons license and therefore has been used to provide translations in this project (see CC-CEDICT, UNIHAN).

[10] The full list of the terms is available on the project site. The list of words that are present in all twenty-four histories would have been much shorter; therefore, the current list is a compromise between the full list of gender-specific terms in all

source, where there are more male actors, and 2) more developed nomenclature of male official titles in the dynastic histories. Although male terms dominate female terms in the curated list, it is the largest gender list for Classical Chinese and serves as a basis for future work.

| 女 | 女 | nǚ | female; woman; daughter |
| 妻 | 妻 | qī | wife |
| 妇 | 婦 | fù | woman; old variant of 妇 |

Table 2: Sample lines from the gender-specific terms.

## 4. Methodology

This work focuses on a simple approach among many possible ways of performing gender analysis. One way of studying would be to create word vectors, based on counting words in a close context of a term and then comparing these vectors and establishing similarity between gender terms. However, comparing the similarities of gender-specific terms or even groups of them was not included in the goals of this study. Rather, this study is aimed at exploring linguistic context of the two groups (male and female) of gender-specific terms. It does so by creating joint context vocabulary for all terms and a co-occurrence matrix (called "the synoptic co-occurrence matrix"[11]), where the target terms are the columns and the context terms are the rows, and synoptic word vectors could be considered as the columns.

This study regards textual and syntactical units, such as sentences and paragraphs (passages), as the two basic units of analysis. The semantic scopes of these syntactic and textual structures are different. It could be expected that the context terms for sentences (that are not common and functional words) are related to collocation and the general structure of the meaning of the target terms. The context terms for paragraphs would be wider in semantic scope since they may describe a discourse topic.

This study also explores semantic analysis of individual terms by focusing on context terms of the gender terms. The authors extract "focus corpora" for a given term, from all sentences where this term is present[12]. Then, the meaning of the term is explored under both topic model and keyword analysis.

## 5. Experiments and Results

### 5.1. Evidence for Synoptic Context of the Gender-Specific Target Terms

The first experiment constructed a synoptic co-occurrence matrix of context terms of male and female terms in the scope of sentences and paragraphs. The tables could be considered count-based word vector table (columns are vectors), where the vocabulary consists of context terms of all gender-specific terms. The total number of all context terms for all target terms in all histories is over 15000 (for sentences); however, with the cutoff values for the context terms, a minimum of 10 entries for the corpus and a minimum of 5 target terms, stop-words included, there are about 6700 sentence-based context terms and 9400 paragraph-based context terms[13]. Several observations can be made:

a) Many context terms are shared by male and female gender-specific terms. Most context words of female terms are shared with male terms. And even though there are many more male terms, a small number of them do not have any shared context terms with female terms. For sentences, there are only about 600 (or about 10% context terms), and those terms are comparatively rare characters, with rarely more than 30 entries in the entire corpus.

b) An analysis of matrix supports grouping target terms according to their distributional features. There are a few words in the target terms that could co-occur with at least 80% of context terms, e.g., 女, 母, 妻, 婦 in female terms and 王, 子, 公, 侯, 君, 臣, 兵, 帝, 士, 父, 孫 in male terms. These terms could be called "star terms" because they are "connected" to most of the context terms, as well as between themselves (as shown later). Then, there are middle-range terms, which are connected to about 50% of the context terms, and, finally, there are low-connected terms.

c) About 8% of the context terms are connected to over 95 target terms for sentences and more than 2000 such terms (around 20%) for paragraphs[14]. Therefore, at the paragraph level, there is a large group of characters that could be in the same context for male and female terms.

It is straightforward to see why context characters for male and female terms overlap considerably, especially for paragraphs. For most paragraphs, where there is at least one female term, there is also at least one male term (this is also true for sentences, but in a lesser degree). Of more than 280000 paragraphs in the corpus[15], 31079 contain at least one female term, and 172909 contain at least one male term; 29371 paragraphs contain at least one male and one female terms, so the number of paragraphs containing female terms and not having male terms is about 5.5%, i.e., if a paragraph topic includes a female actor (designated by gender-specific term, e.g., "wife"), there will almost always be a male actor (e.g., "husband"). However, if a paragraph contains some male terms, e.g., *wang* (king), it would often not contain any female terms.

It is thus not straightforward to establish differences between male and female terms as groups using co-distributional words in this corpus. Female context terms would be subsumed by the male context terms. However, male terms have some semantic space free of female terms, i.e., topically, these paragraphs are not related to

---

any family or other female-related matters[16]. This is one of the forms in which male terms "dominate" female terms in the dynastic histories semantic space.

## 5.2. Evidence for Diachronic Change of Context Terms of the Gender-Specific Target Terms

In the previous section, the study of context terms was conducted on the whole corpus, without diachronic stratification. Having a diachronic corpus of such scale, it is logical to obtain evidence of diachronic changes in the context terms' co-distribution with the target terms.

Additional files were created for this goal[17]. Each row contains pairs of target terms with their context terms, with absolute numbers for each dynastic history in chronological succession. They only contain context terms that have more entries than the cutoff value (four, in this case) at least for one dynasty. The files with normalized numbers contain the same pairs, but with normalized numbers (divided by the dynasty history size in characters). Normalized data allow tracking change for a context character, in combination with a specific target of determining whether its usage is on the rise or decline, i.e., diachronic change in its usage with the context term.

Columns in these tables are similar to the "meaning vectors" used by Xu and Kemp (2015); however, in the current study, instead of measuring target vectors divergence, the authors explored diachronic change of usage of context terms. If for a specific target term, there are many context terms that are either on the rise or decline, it may be an indication of a semantic change in the usage of the given target term.

The direction of change has been identified by linear regression, through normalized values of pairs' entries, on the Y-axis and corpus documents in historical order of writing on the X-axis. The distance between sources is uniform (there is no presumption about the character of temporal change of Classical Chinese, but it definitely would not be linear[18]).

The slope of the linear regression line can indicate the direction of change. In this study, +1.5 and −1.5 were accepted as criteria of change. Pairs that demonstrate slopes more than 1.5 are considered to be on the rise (it will be marked as up); pairs that have slope less than −1.5 are considered to be on the decline (it will be marked as down). The pairs, which have slope in between −1.5 and −1 and 1 and 1.5, are considered to be of "undefined" type. The pairs with slope in the interval of −0.5 and +0.5 are considered to be "neutral," i.e., no definitive change.

The results for paragraphs and sentences are presented in diachronic normalized files in the form described above (the normalized values of pairs' entries are multiplied by factor of 100000 and rounded). There are 250676 lines for paragraphs and 102145 lines for sentences.

It is noted that most pairs belong either to neutral or undefined type. To estimate the directionality of the remaining pairs, they are collected into respective diachronic normalized files on the project site. These files contain 1955 pairs for paragraphs and 210 pairs for sentences.

The analysis of the diachronic change in context and target terms co-occurrence reveals that a very small portion of target–context word/character pairs exhibits considerable change. Most target–context co-occurrence pairs are quite stable over 1500 years, which could be considered as a confirmation of existing, in classic philology, thesis about grammatical and vocabulary stability of Literary Chinese.

## 5.3. Topic Modeling and Keyword Analysis of Gender-Specific Target Terms

The final experiment was a semantic analysis of individual target terms. Two methods have been applied to the corpus in this experiment. Both methods employ the "focus corpus" approach, where a focus corpus is created for each target term based on the passages (within the context window) where the target term is found[19].

The method of clustering passages of a target word to enrich semantic context of a document has been popularized in word sense disambiguation/induction research since the beginning of this century[20], in work by Bordag (2006) (who suggested a sentence-length window) and then in work by Brody and Lapata (e.g., Brody and Lapata, 2009), who, following Caj et al. (2007), used topic modeling.

Using topic modeling for identifying word senses through clustered contextual passages for target words was elaborated by Lau et al. (2012) and Cook et al. (2014), who created focus corpora on the basis of three-sentence context window, where a target term appears in the middle sentence. They conducted a topic analysis, using a hierarchical version (HDP) of Latent Dirichlet Allocation (LDA), to detect novel senses. In their study (Cook et al., 2014), they also implemented keyword analysis (KA) for target terms but yielded only limited use. This study uses similar methods, namely, latent semantic based topic analysis (LSI), which resembles LDA and HDP) and KA[21].

The approach was to apply topic-based analysis (Blei et al., 2003) to a focus corpus, consisting of one-sentence passages that contain the target characters[22]. As an illustration, two representative gendered words from the gender list were chosen: *nan* (man) and *nv* (woman). The focus corpora for these words (based on one sentence passage) have been created, and LDA and KA have been applied to them[23]. The raw output for a run of the program for five topics for LDA is presented in the Table 3.

---

[16] For example, the topic of a paragraph could be an activity of the king (*wang*), and this activity will not include any female actors.

[17] See specific file names in the README file on the OSF project site.

[18] It would rather be expected to be synchronous with spoken Chinese evolution, which is not linear. However, the evolution of Classical Chinese is not a well-researched area.

[19] These passages are considered to be "documents' of the corpus.

[20] This method of aggregating short passages for topic modelling is also not unlike the methods of topic modelling, applied for study of Twitter, etc., corpora. See Hong and Davidson (2010).

[21] Following the standard methods described by Baron et al., (2009) and Scott (2010).

[22] Cook et al. use three-sentence passages for creating a focus corpus. This is a length in between average paragraphs and sentences in their study. However, considering the conciseness of classical Chinese, one sentence in this language could be longer in English.

Many topic characters appear in the topics, and topics are, actually, very close[24]. Because this study is less concerned with specific senses of a word and more concerned with meaning in its generality, topic characters were merged (see Table 4) and redundant characters were removed. Although the result still contains "noise characters" it could be considered an aggregate of senses of the target characters. The merged results are shown in the Table 4 (sorted alphabetically).

The topic modeling correctly identifies the historical characters of the document (official titles, names of kingdoms) and includes some gender-specific terms (see Table 3). For nan character, some topic characters describe societal and political roles of a man (wang (king), gong (Duke), guo (state), xiao (filial piety); others relate to his familial role (nv (woman), fu (father)). For nv character, there are many related male terms (as women's stories are mainly related to their male partners' biographies), such as di (emperor), wang (king), hou (marquis), gong (duke), jun (ruler) – even more so than for nan. There are also a few female gender-specific terms, like hou (empress), mu (mother), qi (wife).

| LDA topics for *nv* | LDA topics for *nan* |
|---|---|
| 子為人王太太生后大長氏<br>為子后太王男人侯生大<br>王為秦子公無取太婦人<br>為公人翏子生兒齊無王<br>子為王公秦母婦太齊無 | 氏子令人日王年城楚太<br>為子人女長王太公帝楚<br>生子人為王女夫后幸<br>女生子無制為人衛梁禮<br>女子人王無國霸長別足 |

Table 3: Topics retrieved from the focus corpus of nv (woman) and nan (man).

An alternative analysis was performed using KA method. Keyword analysis is a popular modern method of content analysis of corpora. It is rooted in Firth (1957) and Williams (2014) and introduced by Mike Scott in his WordSmith software package[25]. KA involves the comparison of word frequencies in a focus corpus and a reference corpus. As such, "keywords are those whose frequency (or infrequency) in a document or corpus is statistically significant, when compared to the standards set by a reference corpus" (Bondi, 2010). The significance of a word as a potential keyword in the KA is measured by its "keyness score."

There are various measures of keyness in the implementation keyword extraction and ranking. This study considers only two main measures: log-likelihood (LL or G2) and chi-square (CHI2 or just CHI). Depending on the measure of keyness, keyword scores may be positive and negative. Positive keywords can be defined as "comparatively overused" words in comparison with word use in the reference corpus, and negative keywords would be "comparatively underused."

The main reason for analyzing corpus documents using keywords is the presumption that they express "aboutness," i.e., they allow understanding of content, based on automatic extraction of frequent words[26].

| Topics for nan (man) | Topics for nv (woman) |
|---|---|
| 人 rén man | 氏 shì clan name |
| 侯 hóu title | 人 rén man |
| 兒 ér child | 令 lìng command |
| 公 gōng title gong | 公 gōng Duke |
| 取 qu take | 別 bié to separate |
| 后 hòu empress | 制 zhì to regulate |
| 大 dà big | 后 hòu empress |
| 太 tài greatest | 國 guo state |
| 婦 fù woman | 城 chéng city |
| 子 zǐ son | 太 tài greatest |
| 母 mǔ mother | 夫 fū husband |
| 氏 shì clan name | 女 nǚ woman |
| 為 wèi do | 子 zǐ son |
| 無 wú not to have 王 | 帝 dì emperor |
| 王 wàng king | 年 nian year |
| 生 shēng give birth | 幸 xìng fortunate |
| 男 nán man (male) | 日 rì day |
| 秦 qín state Qin | 梁 liáng state Liang |
| 長 cháng grow | 楚 chu state Chu |
| 魯 lǔ state Lu | 為 wèi do |
| 齊 qí state Qì | 無 wú not to have |
| | 王 wàng king |
| | 生 shēng give birth |
| | 禮 li ritual |
| | 衛 wèi guard |
| | 足 zú foot |
| | 長 chang grow |
| | 霸 bà feudal chief |

Table 4: Merged LDA sentence topic characters for *nan* and *nv*.

The log-likelihood score (LL or G2) in this study was calculated following the formula, suggested by Paul Rayson (see Rayson in Resources):

$$G2 = 2*((a*\ln (a/E1)) + (b*\ln (b/E2)))[27]$$

---

| | Corp. 1 | Corp. 2 | Total |
|---|---|---|---|
| Freq. of feature | a | b | a+b |
| Freq. of feature not occurring | c | d | c+d |
| Total | a+c | b+d | N=a+b+c+d |

Table 5: Contingency table for the CHI2 test

There are definitely more characters, related to gender context in the KA list. For instance, in the list for *nv* (woman), there are such terms as *fei* (concubine), *ji* (prostitute), *qi* (wife), *qie* (concubine), *fu* (woman), and some terms for marriage, that are not present in the topics. CHI2 score for keywords was calculated using the contingency table and the formula from (Baron et al. 2009), see Table 5.

$$X2 = N(ad-bc)2/(a+b)(c+d)(a+c)(b+d)$$

Similar to the topic analysis, focus corpora were created for target words based on one-sentence context windows (these sentences were removed from the main corpus, which became the reference corpus). The results of the output (alphabetically sorted top thirty characters) are presented in Table 6.

Table 7 summarizes a comparison between topic modeling and keyword analysis. Although there are some overlapping terms in LDA and KA CHI2 lists (e.g., *wang* (king)), there are more differences. Not only there are more terms in the KA output that are found on the gender-specific list, but there is also better fit from the KA to the term's semantics. For example, for woman, on the LDA list, only three terms are from female gender-specific group, while on the KA list there are nine such terms. The KA list of terms offers a better description of women's roles in the traditional Chinese society, while the LDA list instead indicates historical actors to whom women are attached.

It should be recalled, that, from the beginning, topic modeling was not about the "content" of the corpus. If the focus corpus for nv could be defined as consisting of historical passages about women, then topic modeling could be described as extracting its "historical topicality", while KA method, in agreement with how it is defined by its developer, is rather extracting this "aboutness", i.e., the content of the document.

The words, obtained through keyword analysis of focus corpora, can be useful for creating a semantic framework (or a template) of the meaning of the term, i.e., keyword analysis of the focus corpora can be useful for the automatic creation of meaning templates for terms. The list of gender-specific terms can be used as a criterion for projecting gender character of the term in question automatically.

| Nan (man) | Nv (woman) |
|---|---|
| 男 nán man | 主 zhǔ master |
| 伯 bó title bo | 人 rén man |
| 夫 fū husband | 公 gōng duke |
| 女 nǚ woman | 后 hòu empress |
| 好 yú fair | 女 nǚ woman |
| 妻 qī wife | 妃 fēi concubine |
| 姬 jī concubine | 妓 jì prostitute |
| 婕 jié handsome | 妹 mèi younger sister |
| 婚 hūn marry | 妻 qī wife |
| 婦 fù woman | 妾 qiè concubine |
| 嫁 jià marry | 姊 zǐ older sister; |
| 子 zǐ son | 娉 pīng graceful |
| 孕 yùn pregnant | 娶 qǔ marry |
| 封 fēng title | 婚 hūn marry |
| 戶 hù household | 婦 fù woman; |
| 爵 jué vessel | 婢 bi maid |
| 王 wàng king | 壻 xù son-in-law |
| 生 shēng life | 娲 Was surname Wa |
| 產 chǎn to give birth | 嫁 jià marry |
| 癃 lóng infirmity | 嬪 pín imperial concubine |
| 笄 jī hairpin | 子 zǐ son |
| 級 jí rank | 州 zhou province |
| 縣 xiàn county | 母 mǔ mother; |
| 袋 dài bag | 氏 shì clan name |
| 裸 luǒ naked | 淫 yín excess |
| 賜 cì bestow | 男 nán man |
| 邑 yì city | 直 zhí straight |
| 陽 yáng positive | 織 zhī weave |
| 鬘 zhuā dress for hair | 適 kuò proper |
| 鰥 guān widower | 駙 fù prince-consort |

Table 6: Top 30 keyword analysis (CHI2) characters for *nan* (man) and *nv* (woman).

## 6. Discussion and Conclusion

This study contributes to the community the new corpus of Classical Chinese (CCDH), on the basis of open-source dynastic histories, covered by Creative Commons license. The datasets are free for downloading and refined processing (e.g., POS marking) by the public. This corpus is unique in that it supports both synchronic and diachronic studies of Classical Chinese, where there is a dearth of free available and licensed corpora[28].

The second contribution of the study is to offer a case study of how this corpus could be used, in terms of gender analysis. Gender analysis is not yet a developed area of research in Classical Chinese, so the authors had to create a novel list of gender-specific terms. The study creates a co-occurrence matrix of target terms from the gender-specific list and their context terms. The list of context terms ("synoptic vocabulary") underlines contextual

---

"a+b" will be the total number of a character in both corpora, and "c+d" is the number of all characters in both corpora. In these terms, expected values E1 (for Corpus 1) and E2 (for Corpus 2) will be E1 = c*(a+b) /(c+d) and E2 = d*(a+b)] / (c+d).

relations to the target terms. It is found that most target terms share context, but the male terms have larger synoptic vocabularies (i.e., more diverse context).

| LDA nan | KA nan | LDA nv | KA nv |
|---|---|---|---|
| 人 man | 男 man | 人 man | 主 master |
| 侯 title | 伯 title bo | 公 Duke | 人 man |
| 兒 child | 夫 husband | 后 empress | 公 duke |
| 公 gōng | 女 woman | 夫 husband | 后 empress |
| 后 empress | 妻 wife | 女 woman | 女 woman |
| 婦 woman | 姬 concubine | 子 son | 妃 concubine |
| 子 son | 婦 woman | 帝 emperor | 妓 prostitute |
| 母 mother | 子 son | 王 king | 妹 sister |
| 王 king | 王 king | 霸 feudal chief | 妻 wife |
| 男 man | 鰥 widower | | 妾 concubine |
| | | | 姊 older sister |
| | | | 婦 woman |
| | | | 婢 maid |
| | | | 壻 son-in-law |
| | | | 嬪 concubine |
| | | | 子 son |
| | | | 母 mother |
| | | | 駙 consort |

Table 7: Comparison of LDA and KA terms, found on the gender-specific list of terms.

The diachronic analysis of context terms in the synoptic vocabularies reveals that these vocabularies are relatively stable, i.e., not many pairs of context-target terms display substantial change (or considerable increase and decrease in frequency over time). This study opens up opportunities for future inquiries into semantic change and the historical lexicon in Classical Chinese.

## 7. Acknowledgements

We would like to thank the anonymous reviewers and Ella Rabinovich for their helpful comments and suggestions.

## 8. Bibliographical References

## 9. Language Resource References